\documentclass[conference]{IEEEtran}
\IEEEoverridecommandlockouts
\usepackage[dvips]{graphicx}
\usepackage{algorithm}
\usepackage{algorithmic}
\usepackage{psfrag}

\usepackage{bm}
\usepackage{amssymb,mathtools}
\usepackage{amsfonts}
\usepackage{amsmath,amssymb}
\usepackage{subfigure}
\usepackage{amsthm}
\usepackage{cite} 
\usepackage[ancient]{flushend}
\usepackage[dvipsnames]{xcolor}
\usepackage{url}
\usepackage[hyperfootnotes=false]{hyperref}
\usepackage{graphicx}
\usepackage{upgreek}
\usepackage{yfonts}
\usepackage{cleveref}
\usepackage{caption}
\usepackage{amsmath}
\usepackage{cite}

\begin{document}

\title{Resource-Aware Asynchronous Online Federated Learning for Nonlinear Regression
}

\author{Francois Gauthier$^{\star}$, Vinay Chakravarthi Gogineni$^{\star}$, Stefan Werner$^{\star}$, Yih-Fang Huang$^{\dagger}$, Anthony Kuh$^{\ddagger}$  \thanks{This work was supported by the Research Council of Norway.}\\

$^{\star}$Dept. of Electronic Systems, Norwegian University of Science and Technology, Norway\\ $^{\dagger}$Dept. of Electrical Engineering, University of Notre Dame, Notre Dame, IN, USA\\ $^{\ddagger}$Dept. of Electrical and Computer Engineering, University of Hawaii, Hawaii, USA\\E-mails: \{francois.gauthier, vinay.gogineni, stefan.werner\}@ntnu.no, huang@nd.edu, kuh@hawaii.edu
}
\makeatletter
\newcommand{\linebreakand}{%
  \end{@IEEEauthorhalign}
  \hfill\mbox{}\par
  \mbox{}\hfill\begin{@IEEEauthorhalign}
}
\makeatother

\newcommand{\Lim}[1]{\raisebox{0.5ex}{\scalebox{0.8}{$\displaystyle \lim_{#1}\;$}}}

\maketitle

\begin{abstract}
Many assumptions in the federated learning literature present a best-case scenario that can not be satisfied in most real-world applications. An asynchronous setting reflects the realistic environment in which federated learning methods  must be able to operate reliably. Besides varying amounts of non-IID data at participants, the asynchronous setting models heterogeneous client participation due to available computational power and battery constraints and also accounts for delayed communications between clients and the server. To reduce the communication overhead associated with asynchronous online federated learning (ASO-Fed), we use the principles of partial-sharing-based communication. In this manner, we reduce the communication load of the participants and, therefore, render participation in the learning task more accessible. We prove the convergence of the proposed ASO-Fed and provide simulations to analyze its behavior further. The simulations reveal that, in the asynchronous setting, it is possible to achieve the same convergence as the federated stochastic gradient (Online-FedSGD) while reducing the communication tenfold. 
\end{abstract}

\section{Introduction}
A vast amount of data is available on distributed devices, including edge devices. This motivates the development of distributed learning methods operating on those devices, and coping with edge devices. Federated learning (FL) is designed with this task in mind, as it provides an adaptive large-scale collaborative learning framework. In FL, a server aggregates information received from devices called clients to train a global model; the clients do not share any private data with the server, only their local model parameters \cite{origin,FedOverview}. Amongst the features that make federated learning (FL) stand out from typical distributed learning are the assumptions of uneven data distribution and statistical heterogeneity \cite{Noniid}. 

There are, however, further complications when learning on edge devices that are not considered in most FL implementations \cite{FedAVG, Accelerated, FedOverview, FedAvgCV, Compression, Energy, FLCompress1,FLCompress2,FLCompress3}. In a realistic setting, clients cannot be expected to have the same participation frequency, e.g., because of battery constraints, channel availability, or concurrent solicitations\cite{Async,Async2, AsyncOnline}. Furthermore, clients may become unavailable for a certain period during the learning process, i.e., some clients are temporarily out of order or not reachable by the server \cite{Async, Async2}. Physical constraints may also introduce delays in the communication between clients and the server \cite{Async2, AsyncOnline, AsyncTier}. These constraints, frequently occurring in practice, impair the efficiency of existing FL methods and make the development of FL methods tailored for an asynchronous environment challenging  \cite{Async, AsyncOnline, AsyncTier, Async2, CommAsync,Asyncstuf}.

Besides, the energy toll taken by FL methods on clients can sometimes be prohibitive, and methods are being developed to address this issue \cite{Energy, AsyncTier}. Notably, the communication of model parameters to and from the server accounts for a substantial portion of the power consumption in FL. Hence, it is crucial to reduce this communication cost from the client's perspective \cite{AsyncTier, Energy, FedAVG, CommAsync}. Furthermore, in asynchronous settings, efficiency in communication is essential for reducing energy and bandwidth thresholds at which devices choose to participate, hence increasing participation rates, and thus maximizing the benefits of the FL framework. Therefore, clients would benefit from an FL approach that ensures satisfactory learning in an asynchronous environment while imposing a minimal communication burden. 

A considerable amount of research has been undertaken on both communication-efficient FL \cite{FedAVG, Compression,Vinay, DblCompress, RandomMask,FLCompress1,FLCompress2,FLCompress3} and asynchronous FL \cite{Async, AsyncOnline, Async2, Asyncstuf}; however, there is little work combining the two aspects into one. The works in \cite{AsyncTier,DblCompress} reduced the communication overhead via compressed updates on the client-side, but they did not address communication needs on the server-side. Aside from the accuracy cost associated with this projection method, it also adds an extra computational burden, which is not appealing for low-battery clients. Moreover, the work in \cite{DblCompress} did not consider asynchronous settings. Although the work in \cite{CommAsync} reduced the communication load of the clients, it is specific to deep neural networks and does not provide a mathematical analysis of the presented results; in addition, the asynchronous setting considered do not include communication delays. The classical federated averaging (Fed-Avg) \cite{FedAVG} reduced the communication in FL by selecting a subset of the clients to participate at each iteration. However, because some clients may participate sporadically in the asynchronous setting, we do not intend to discard any participation by sub-sampling the available clients. Another option, explored until recently only in distributed learning, is the partial-sharing of the model parameters \cite{PartialSharing}. The partial-sharing-based online FL (PSO-Fed) \cite{Vinay} introduces partial-sharing  in FL, but only in an ideal setting.


This paper extends the work and analysis on partial-sharing-based communication to the asynchronous online FL. The developed method presents the advantages of being feasible in a realistic environment and reducing the computational load of participants.
The proposed partial-sharing asynchronous online federated learning (PAO-Fed) algorithm oversees the collaborative estimation of a continuous nonlinear model represented on a random Fourier feature (RFF) space where only subsets of the RFF representation of the model are exchanged between the server and the clients. We prove that PAO-Fed converges even in a setting where client participation is random, and communication links suffer from delays. Lastly, we provide numerical results to compare PAO-Fed with existing methods in various asynchronous settings.


\section{Preliminaries and Problem Formulation}

\subsection{Federated learning}

We consider a global server linked to $K$ geographically distributed devices, referred to as clients. The set of clients is denoted $\mathcal{K}$. The data of the network is distributed over those client, the local data of client $k \in \mathcal{K}$ is denoted ${\bf X}_k$. We consider that the entire data is not available at the beginning of the learning process but instead becomes available progressively. We denote the data available at client $k$ at global iteration $n$ by ${\bf x}_{k,n}$ and the corresponding desired output by ${\bf y}_{k,n}$, their relation is described as
\begin{align}
    y_{k,n} = f({\bf x}_{k,n}) + \eta_{k,n},
\end{align}
\noindent where $f(\cdot)$ is a continuous nonlinear model that we want to estimate, and $\eta_{k,n}$ is the observation noise. For the server to estimate $f(\cdot)$, the clients will share model parameters learned from their local data to the server. We denote ${\bf w}_n$ and ${\bf w}_{k,n}$ the server and local (at client $k$) model parameter vectors, they are linear representations of the nonlinear model $f(\cdot)$ in the random Fourier feature (RFF) space of dimension $D$. The objective functions at the server and at the clients are given, as in \cite{RFFKLMSB,RFFKLMS}, by:
\begin{align}
    \mathcal{J}({\bf w}_n) &=  \frac{1}{K} \sum_{k \in \mathcal{K}} \mathcal{J}_k({\bf w}_n) \\
    \mathcal{J}_k({\bf w}_{k,n}) &= \mathbb{E} [|y_{k,n} - \hat{y}_{k,n} |^2 ], \notag
\end{align}
\noindent with $\hat{y}_{k,n} = {\bf w}_{k,n}^{\mathsf{T}} {\bf z}_{k,n}$, ${\bf z}_{k,n}$ being the mapping of ${\bf x}_{k,n}$ into the RFF space.

\subsection{Online-Fed}

At each global iterations, the server selects a subset of clients $\mathcal{K}_n \in \mathcal{K}$ to participate in the learning. It shares the global model ${\bf w}_{n}$ with the clients in $\mathcal{K}_n$ for them to perform the following local update step: 
\begin{align}
   {\bf w}_{k,n+1} = {\bf w}_{n} + \mu \epsilon_{k,n} {\bf z}_{k,n},
\end{align}
\noindent where $\mu$ is the learning rate and $\epsilon_{k,n} = y_{k,n} - {\bf w}_{n}^{\mathsf{T}} {\bf z}_{k,n}$. The clients proceed to share their new models with the server who aggregates them as:
\begin{align}
    {\bf w}_{n+1} = \frac{1}{| \mathcal{K}_n |} \sum_{k \in \mathcal{K}_n} {\bf w}_{k,n+1}.
\end{align}

In the particular case where, at each global iteration, $\mathcal{K}_n = \mathcal{K}$, that is, all the clients are always selected to participate, we denote the algorithm Online-FedSGD.

\subsection{PSO-Fed}
The PSO-Fed algorithm, defined in \cite{Vinay}, reduces the communication overhead of the Online-Fed algorithm without compromising the accuracy. In contrast, projection-based models see, e.g., \cite{AsyncTier,DblCompress}, require increased computations while suffering some loss in accuracy. The PSO-Fed algorithm uses partial sharing-based communication (see \cite{PartialSharing}) and transfers only a subset of the model at each iteration. The model subsets shared by the server (to client) and client $k$ (to server), at iteration $n$, are determined by selection matrices ${\bf M}_{k,n}$ and ${\bf S}_{k,n}$, respectively. In particular, ${\bf M}_{k,n}$ and ${\bf S}_{k,n}$ are diagonal matrices with diagonal values of either $0$ or $1$, where the locations of the latter specify the design parameters to share. 

In addition, the PSO-Fed allows clients to perform local updates at arbitrary time instants, regardless of whether they take part in the global aggregation step. Specifically, if client $k$ receives new data at an iteration $n$ when it is not assisting in the global update, it updates its model as:
\begin{align}
    \label{LocalUpdate}
   {\bf w}_{k,n+1} = {\bf w}_{k,n} + \mu \epsilon_{k,n} {\bf z}_{k,n},
\end{align}
\noindent where $\epsilon_{k,n} = y_{k,n} - {\bf w}_{k,n}^{\mathsf{T}} {\bf z}_{k,n}$.

By combining scheduling and partial-sharing, the PSO-Fed algorithm can significantly reduce the amount of communication. Furthermore, it converges rapidly thanks to the local update performed independently by the clients, allowing them to refine their model prior to sharing it when selected to participate. The aggregation step at the server is given by:
\begin{align}
    {\bf w}_{n+1} &= {\bf w}_n + \frac{1}{|\mathcal{K}_{n}|} \sum_{k \in \mathcal{K}_{n}} {\bf S}_{k,n} ({\bf w}_{k,n} - {\bf w}_n).
\end{align}

The main feature of the PSO-Fed algorithm is the considerable cut in communication overhead; however, it does not incorporate knowledge of the network environment or client resources. For example, when performing federated learning in real-time, clients may be unavailable for various reasons, e.g., limited battery power and channel availability. In addition, poor connection or sub-optimal communication channels may delay exchanged messages. Delays can also arise from some clients with limited computational resources that struggle to perform the task in time (so-called straggler-clients). Therefore, for a successful operation in a realistic environment, an online federated learning method needs to handle time-varying client participation and weigh the importance of delayed measurements. To that end, in this work, we modify the PSO-Fed algorithm to handle such an asynchronous setting. As we will see in the following, many choices made for the PSO-Fed algorithm in a perfect FL setting are not the best in the asynchronous setting.

\section{Proposed method}

\subsection{Asynchronous setting}

We consider an asynchronous setting where clients have access to uneven amounts of non-IID data and have various availability, meaning some will participate more often than others to the learning. In addition, clients may become unavailable for several iterations during the learning process. Furthermore, we expect communication from the clients to the server to be subject to delays. 

The consequence of the introduced delays is that not all updates from clients will arrive at the server simultaneously. An update sent by a client at time $n-l$ and delayed for $l$ iterations, will be received by the server at time $n$. We denote by $\mathcal{K}_{n,l}$ the set of all the clients who sent an update at iteration $n-l$ that reached the server at iteration $n$, the subscript $l$ corresponds to the number of iterations during which the update was delayed. Further, we define the set $\mathcal{K}_n = \sum_{l=0}^{\infty} \mathcal{K}_{n,l}$ of all the clients who sent an update that arrived at the server at iteration $n$. When receiving a delayed update from a client, a decision should be made whether to use this update, as it may be outdated. To improve the learning accuracy of the algorithm, we propose an aggregation mechanism, inspired by \cite{AoI}, where weights are given to received local models according to how recent they are.

\subsection{PAO-Fed}

In this setting, we chose not to sub-sample the available clients, some of whom might be rarely available. Instead, we will rely on partial-sharing to reduce the communication overhead in the asynchronous FL setting. At iteration $n$, the server will share a subset of its model with all the available clients. Further, the clients will share a subset of their models to the server, but although this reply is sent at iteration $n$, it may arrive to the server later. Clients receiving new data perform the local update step \eqref{LocalUpdate} if they are not available, allowing them to communicate a refined model later on. Performing this local update is a trivial computation for most devices and does not require communication with the server. 

At global iteration $n$, the server shares a subset of its model, ${\bf w}_n$, to all the available clients, the selection matrices ${\bf M}_{k,n}$ dictates which subset goes to which client. Each available client $k$ receives ${\bf M}_{k,n} {\bf w}_n$ and updates its local model as
\begin{align}
\label{ClientUpdateSelected}
    {\bf w}_{k,n+1} &= {\bf M}_{k,n} {\bf w}_n + ({\bf I} - {\bf M}_{k,n}) {\bf w}_{k,n} + \mu e_{k,n} {\bf z}_{k,n}, 
\end{align}
where the error $e_{k,n}$  is given by
\begin{align}
\label{Error_available}
    e_{k,n} &=  y_{k,n} - ({\bf M}_{k,n} {\bf w}_n + ({ \bf I} - {\bf M}_{k,n}) {\bf w}_{k,n} )^{\text{T}} {\bf z}_{k,n}.
\end{align}
Then, the available clients communicate a portion of their updated local model, i.e., ${\bf S}_{k,n+1} {\bf w}_{k,n+1}$, to the server; this communication may be delayed. At the server, we consider the set of clients $\mathcal{K}_n$ whose updates arrived at the server at time $n$. We decompose this set according to the number of iterations during which the updates were delayed and update the server model as:
\begin{align}
\label{ServerUpdate}
    {\bf w}_{n+1} &= {\bf w}_n + \sum_{l = 0}^{\infty} \frac{\alpha_l}{|\mathcal{K}_{n,l}|} \sum_{k \in \mathcal{K}_{n,l}} {\bf S}_{k,n-l} ({\bf w}_{k,n-l} - {\bf w}_n), 
\end{align}
while omitting any empty set $\mathcal{K}_{n,l}$, where $\alpha_l$ is the weight given to the updates delayed by $l$ iterations. We denote by $l_{max}$ the maximum delay after which $\alpha_l = 0$. The weight $\alpha_0$ given to the updates that are not delayed is $1$, we will see in the simulation section the importance having $\alpha_l < 1$ for $l>1$. The resulting algorithm is presented in Algorithm \ref{AsyncAlgo}.

We note that in the eventuality where several clients in $\mathcal{K}_{n}$ update the same model parameter, only the most recent updates are considered, the selection matrices of the remaining updates are adjusted accordingly prior to computing \eqref{ServerUpdate}.

\begin{algorithm}[h!]
\caption{PAO-Fed}
\label{AsyncAlgo}
\begin{algorithmic}[1]
    \STATE Initialization: ${\bf w}_0$ and ${\bf w}_{k,0}, k \in \mathcal{K}$ set to $\boldsymbol{0}$
    \STATE Procedure at Local client $k$
    \FOR{ global iteration $n=1,2,\hdots,N$}
        \IF{Client $k$ receives new data at time $n$}
            \IF{$k$ is available}
                \STATE Receive ${\bf M}_{k,n} {\bf w}_{n}$ from the server.
                \STATE Compute ${\bf w}_{k,n+1}$ as in \eqref{ClientUpdateSelected} .
                \STATE Share ${\bf S}_{k,n+1} {\bf w}_{k,n+1}$ to the server.
            \ELSE
                \STATE Update ${\bf w}_k$ as in \eqref{LocalUpdate}.
            \ENDIF
        \ENDIF
    \ENDFOR
    \STATE Procedure at Central Server
    \FOR{ global iteration $n=1,2,\hdots,N$}
        \STATE Receive client updates from subset $\mathcal{K}_n \subset \mathcal{K}$.
        \STATE Share ${\bf M}_{k,n} {\bf w}_{n}$ with the available clients. 
        \STATE Compute ${\bf w}_{n+1}$ as in \eqref{ServerUpdate}.
	\ENDFOR
\end{algorithmic} 
\end{algorithm}

\section{Convergence analysis}
 
First, we present the update steps of the algorithm in matrix form. Similar to \cite{tcas2}, before proceeding to the analysis, we define the extended model vector ${\bf w}_{e,n}\in \mathbb{R}^{D(1+K (l_{max}+1)) \times 1}$:
\begin{align}
    {\bf w}_{e,n} &= ({\bf w}_{n}^{\mathsf{T}},{\bf w}_{1,n}^{\mathsf{T}}, \hdots, {\bf w}_{K,n}^{\mathsf{T}}, {\bf w}_{1,n}^{\mathsf{T}}, \hdots, {\bf w}_{K,n}^{\mathsf{T}}, \\
    &{\bf w}_{1,n-1}^{\mathsf{T}}, \hdots, {\bf w}_{K,n-1}^{\mathsf{T}}, \hdots, {\bf w}_{1,n-l_{max}}^{\mathsf{T}}, \hdots, {\bf w}_{K,n-l_{max}}^{\mathsf{T}})^{\mathsf{T}}. \notag
\end{align}
Further, we define the local update matrix ${\bf A}_n \in \mathbb{R}^{D (K+1)}$:
\begin{align}
    {\bf A}_n = 
    \begin{bmatrix}
     {\bf I} & \boldsymbol{0}_D & \dotsb & \boldsymbol{0}_D \\
     a_{1,n} {\bf M}_{1,n} & {\bf I} - a_{1,n} {\bf M}_{1,n} &  & \vdots \\
     \vdots & \boldsymbol{0}_D & \ddots & \boldsymbol{0}_D \\
     a_{K,n} {\bf M}_{K,n} & \vdots & & {\bf I} - a_{K,n} {\bf M}_{K,n} \\
     \end{bmatrix},
     \notag
\end{align}
where $a_{k,n} = 1$ if $k \in \mathcal{K}_{n}$ and $0$ otherwise, and extend it into ${\bf A}_{e,n} \in \mathbb{R}^{D(1+K (l_{max}+1))}$ defined as a block matrix with ${\bf A}_n$ on the top left corner and the value of the identity matrix of $\mathbb{R}^{D(1+K (l_{max} + 1))}$ where ${\bf A}_n$ is not defined. The mapping of the available data at global iteration $n$ into the RFF space is given by ${\bf Z}_{n} = \text{blockdiag}\{{\bf z}_{1,n}, \dotsb, {\bf z}_{K,n}\} \in \mathbb{R}^{D K}$, where $\text{blockdiag}\{\cdot\}$ denotes the block diagonalization operator. We define its extended version as ${\bf Z}_{e,n} = (\boldsymbol{0}_{K \times D K}, {\bf Z}_{n}, \boldsymbol{0}_{D K}, \hdots, \boldsymbol{0}_{D K})$. We also define the observation noise, $\boldsymbol{\eta}_{n} = (\eta_{1,n}, \dotsb, \eta_{K,n})^\mathsf{T}$, and its extended version $\boldsymbol{\eta}_{e,n} \in \mathbb{R}^{1+K (l_{max}+1) \times 1}$ defined as $\boldsymbol{\eta}_{n}$ between index $K+1$ and $2K$, and zero otherwise. Finally, we introduce ${\bf w}_{e}^*$, a vector composed of ${\bf w}^*$ concatenated $1+K (l_{max}+1)$ times. We can now express the extended observation vector as:
\begin{align}
    {\bf y}_{e,n} = {\bf Z}_{e,n}^{\mathsf{T}} {\bf w}_{e}^* + \boldsymbol{\eta}_{e,n},
\end{align}
where the values of ${\bf y}_{e,n}$ between index $K+1$ and $2K$ correspond to $(y_{1,n}, \dotsb, y_{K,n})$. Further, we can express the estimation error as:
\begin{align}
    \boldsymbol{e}_{e,n} = {\bf y}_{e,n} - {\bf Z}_{e,n}^\mathsf{T} {\bf A}_{e,n} {\bf w}_{e,n}, \notag
\end{align}
whose values between index $K+1$ and $2K$ correspond to $(e_{1,n}, \dotsb, e_{K,n})$. We see that:
\begin{align}
    \boldsymbol{e}_{e,n} = {\bf Z}_{e,n}^\mathsf{T} ({\bf w}_{e}^* - {\bf A}_{e,n} {\bf w}_{e,n}) + \boldsymbol{\eta}_{e,n}. \notag
\end{align}

We note that performing ${\bf w}_{e,n} = {\bf A}_{e,n} {\bf w}_{e,n} + \mu {\bf Z}_{e,n} \boldsymbol{\epsilon}_{e,n}$ has the effect of updating the first appearance of $({\bf w}_{1,n}, \hdots, {\bf w}_{K,n})$ in ${\bf w}_{e,n}$ to $({\bf w}_{1,n+1}, \hdots, {\bf w}_{K,n+1})$. In order to model the global update step of the algorithm, as well as the change from ${\bf w}_{e,n}$ to ${\bf w}_{e,n+1}$, we introduce the following matrix:
\begin{align}
    {\bf B}_{e,n} &=
    \begin{bmatrix}
     \boldsymbol{B}_{n} & \boldsymbol{B}_{0,n} & \boldsymbol{0}_{D \times DK}  & \boldsymbol{B}_{1,n} & \dotsb & \boldsymbol{B}_{l_{max},n} \\
     \boldsymbol{0}_D & {\bf I}_{D K} & \boldsymbol{0}_{D K} & \dotsb & \dotsb & \boldsymbol{0}_{D K} \\
     \vdots & {\bf I}_{D K} & \boldsymbol{0}_{D K} & \dotsb & \dotsb & \boldsymbol{0}_{D K} \\
     \vdots & \boldsymbol{0}_{D K} & {\bf I}_{D K} & \boldsymbol{0}_{D K} & \dotsb & \boldsymbol{0}_{D K} \\
     \vdots & \vdots & \ddots & \ddots & \ddots & \boldsymbol{0}_{D K}\\
     \boldsymbol{0}_D & \boldsymbol{0}_{D K} & \dotsb & \boldsymbol{0}_{D K} & {\bf I}_{D K} & \boldsymbol{0}_{D K} \\
     \end{bmatrix}
     \notag \\
     \boldsymbol{B}_{n} &= {\bf I} - \sum_{l=0}^{l_{max}} \alpha_l \sum_{k \in \mathcal{K}_{n,l}} \frac{a_{k,n,l} }{|\mathcal{K}_{n,l}|} {\bf S}_{k,n+1-l} \notag \\
     \boldsymbol{B}_{l,n} &= (\frac{\alpha_l a_{1,n,l}}{|\mathcal{K}_{n,l}|} {\bf S}_{1,n+1-l}, \dotsb, \frac{\alpha_l a_{K,n,1}}{|\mathcal{K}_{n,l}|} {\bf S}_{K,n+1-l}). \notag
\end{align}
\noindent where $a_{k,n,l} = 1$ if $k \in \mathcal{K}_{n,l}$ and $0$ otherwise.

We can express both the local and global update steps as:
\begin{align}
\label{Total_update_bigmatrices}
    {\bf w}_{e,n+1} = \boldsymbol{B}_{e,n} ({\bf A}_{e,n} {\bf w}_{e,n} + \mu {\bf Z}_{e,n} \boldsymbol{\epsilon}_{e,n}). 
\end{align}

We will now prove the mean convergence of the PAO-Fed algorithm under the following assumptions:

\noindent \textbf{Assumption 1:} The mapping of the data vectors ${\bf z}_{k,n}, k \in \mathcal{K}$ are drawn at each time step from a WSS multivariate random sequence with correlation matrix $\boldsymbol{R}_k = \mathbb{E}[{\bf z}_{k,n} ({\bf z}_{k,n})^\mathsf{T}]$.

\noindent \textbf{Assumption 2:} The observation noise $\eta_{k,n}$ is assumed to be white, i.i.d., and independent of all input and output data.

\noindent \textbf{Assumption 3:} At each client, the model parameter vector is assumed to be independent of the input data.

\noindent \textbf{Theorem I:} Under assumption 1,2, and 3, $\Tilde{{\bf w}}_{e,n} = {\bf w}_e^* - {\bf w}_{e,n}$ converges to $\boldsymbol{0}$ as $n$ goes to infinity for
\begin{align}
\label{Convergence_condition}
    0 < \mu < \frac{2}{ \max_{\forall k \in \mathcal{K}} \max_{\forall i} \lambda_i(\boldsymbol{R}_k) }.
\end{align}

\begin{proof}
First, we note that by construction we have ${\bf w}_e^* = {\bf B}_{e,n} {\bf w}_e^* = {\bf A}_{e,n} {\bf w}_e^*$ (as all lines in ${\bf B}_{e,n}$ and ${\bf A}_{e,n}$ sum to $1$). Then, using \eqref{Total_update_bigmatrices}, we can recursively express $\Tilde{{\bf w}}_{e,n}$:
\begin{align}
    \Tilde{{\bf w}}_{e,n+1} &= {\bf w}_e^* - {\bf w}_{e,n+1} \\
    &= {\bf w}_e^* - \boldsymbol{B}_{e,n} {\bf A}_{e,n} {\bf w}_{e,n} - \boldsymbol{B}_{e,n} \mu {\bf Z}_{e,n} \boldsymbol{\epsilon}_{e,n} \notag\\
    &= \boldsymbol{B}_{e,n} {\bf A}_{e,n} \Tilde{{\bf w}}_{e,n} - \boldsymbol{B}_{e,n} \mu {\bf Z}_{e,n} \boldsymbol{\eta}_{e,n} \notag \\
    &- \boldsymbol{B}_{e,n} \mu {\bf Z}_{e,n} {\bf Z}_{e,n}^\mathsf{T} ({\bf w}_{e}^* - {\bf A}_{e,n} {\bf w}_{e,n}) \notag \\
    &= \boldsymbol{B}_{e,n} ({\bf I} - \mu {\bf Z}_{e,n} {\bf Z}_{e,n}^\mathsf{T}) {\bf A}_{e,n} \Tilde{{\bf w}}_{e,n} \notag \\
    &- \mu \boldsymbol{B}_{e,n} {\bf Z}_{e,n} \boldsymbol{\eta}_{e,n}. \notag
\end{align}

We now take the expectation of the previous results:
\begin{align}
    \mathbb{E} [\Tilde{{\bf w}}_{e,n+1}] &= \mathbb{E} [\boldsymbol{B}_{e,n}] \mathbb{E} [{\bf I} - \mu {\bf Z}_{e,n} {\bf Z}_{e,n}^\mathsf{T}] \mathbb{E} [{\bf A}_{e,n}] \mathbb{E} [\Tilde{{\bf w}}_{e,n}] \notag \\
    \mathbb{E} [\Tilde{{\bf w}}_{e,n+1}] &= \mathbb{E} [\boldsymbol{B}_{e,n}] ({\bf I} - \mu \boldsymbol{R}_e) \mathbb{E} [{\bf A}_{e,n}] \mathbb{E} [\Tilde{{\bf w}}_{e,n}], \notag
\end{align}
where $\boldsymbol{R}_e = \text{blockdiag}\{ \boldsymbol{0}_{D}, \boldsymbol{R}_1, \boldsymbol{R}_1, \dotsb, \boldsymbol{R}_K, \boldsymbol{0}_{D(K l_{max})} \}$.

Further, we consider the block $ \Tilde{{\bf w}}_{e,n} |_{[D+1,D(K+1)]} = ({\bf w}^* - {\bf w}_{1,n}, \hdots, {\bf w}^* - {\bf w}_{K,n}) $, that is defined as a linear sequence of order $1$ in the normed algebra $\mathbb{R}^{D K}$: $ \mathbb{E} [\Tilde{{\bf w}}_{e,n+1} |_{[D+1,D(K+1)]}] = ({\bf I} - \mu \boldsymbol{R}_{e}|_{[D+1,D(K+1)]}) {\bf A}_{e,n} |_{[D+1,D(K+1)]} \mathbb{E} [\Tilde{{\bf w}}_{e,n} |_{[D+1,D(K+1)]}]$. To prove the convergence of $\mathbb{E} [\Tilde{{\bf w}}_{e,n} |_{[D+1,D(K+1)]}]$ to zero, we use the infinity norm. From the definition of ${\bf A}_{n}$, We have $|| {\bf A}_{n} |_{[D+1,D(K+1)]} ||_{\infty} = 1$. Then the convergence condition reduces to $||\boldsymbol{I} - \mu \boldsymbol{R}_{e}|_{[D+1,D(K+1)]}||_{\infty} < 1$, equivalently,  $|1 - \mu \lambda_i(\boldsymbol{R}_k)| < 1$, $\forall k \in \mathcal{K}, \forall i$, where $\lambda_i(\cdot)$ is the $i^{th}$ eigenvalue of the argument matrix. This leads to the first-order convergence condition given by \eqref{Convergence_condition} as, if $ \Tilde{{\bf w}}_{e,n} |_{[D+1,D(K+1)]} $ converges to zero, then, by construction, $ \Tilde{{\bf w}}_{e,n} $ too.

\end{proof}

\section{Numerical Simulations}

\subsection{Simulation setting}
We model the uneven client participation by giving each client $k \in \mathcal{K}$ a probability $p_{k,n}$ to participate in the learning at global iteration $n$. This probability is $0$ if client $k$ does not receive any new data at that time. The probabilistic nature of the selection enables us to model both the varied availability and potential downtime of the clients. The Bernouilli trial on $p_{k,n}$ dictates if a client is available or not at a given iteration. Further, each communication to the server has a probability $\delta^l, l \in \mathbb{N}$ of being delayed by $l$ global iterations or more; this probability is assumed to be the same for all the clients.

The number of model parameters shared at each step is denoted $m$ and corresponds to the number of nonzero elements in the selection matrices ${\bf M}_{k,n}$ and ${\bf S}_{k,n}$. If $\forall k,n, \: {\bf M}_{k,n} = {\bf S}_{k,n}$, the local updates of the unavailable clients are not being used. In contrast, if $\forall k,n, \: {\bf S}_{k,n} = \text{circshift}({\bf M}_{k,n},m)$ and $\forall k,n, \: {\bf S}_{k,n+1} = \text{circshift}({\bf S}_{k,n},m), {\bf M}_{k,n+1} = \text{circshift}({\bf M}_{k,n},m)$, the local updates are being used as much as possible. The operator $\text{circshift}$ denotes a circular shift. In addition, we call coordinated partial-sharing the specific case where $\forall n,k \neq k', \: {\bf M}_{k,n}={\bf M}_{k',n}, {\bf S}_{k,n} = {\bf S}_{k',n}$; otherwise the partial-sharing is called uncoordinated. In the simulations, we implement uncoordinated partial-sharing with ${\bf M}_{k,n} = \text{circishift}({\bf M}_{1,n}, m k)$ and ${\bf M}_{1,n} = \text{circishift}({\bf M}_{1,0}, m n)$. It is shown in \cite{Vinay} that coordinated outperforms uncoordinated in a perfect FL setting.

We consider an RFF space of dimension $D=200$ and $K = 256$ clients separated into $4$ data groups for which the progressively available training set is of size $500$, $1000$, $1500$, and $2000$, respectively. The clients of each data group are further separated into $4$ availability groups, dictating their probability to participate at each iteration. We consider $2$ different asynchronous settings. In Setting I, the participation probabilities for the availability groups are $0.25$, $0.1$, $0.025$, and $0.005$; and each communication to the server will be delayed by more than $l$ global iterations with probability $\delta^l, 0<l<l_{max}$, with $\delta = 0.2$ and $l_{max} = 10$. Setting II models a harsher environment, availability groups are given the probabilities $0.025$, $0.01$, $0.0025$, and $0.0005$; and communications to the server have a probability $\delta = 0.4$ to be delayed. Further, delays last for more than $l$ global iterations, $l$ taking the values $10 i, 0 \leqslant i \leqslant 6$, with probability $\delta^{\frac{l}{10}}$; $l_{max}$ is set to $60$. This notably implies that, in Setting II, delayed updates have a greater probability to arrive after a non-delayed update coming from the same client.

Unless specified otherwise, every PAO-Fed-based algorithm is set to $m = 4$, and, therefore, reduce the communication load of the algorithm by $98 \%$. We consider seven different versions of the PAO-Fed algorithm. In particular, methods whose names start with PAO-Fed-C use coordinated partial-sharing, while those with PAO-Fed-U use uncoordinated partial-sharing.  Similarly, the methods whose names end with 0 do not use the local update steps, and those whose names end with 1 make full use of those. All the aforementioned methods are set up with $\alpha_l = 1, 0 \leqslant l \leqslant l_{max}$, the method titled PAO-Fed-C2 is identical to PAO-Fed-C1 except for the fact that it is set up with $\alpha_l = 0.2^l, 0 \leqslant l \leqslant l_{max}$. We evaluate the performance of the algorithms with the mean squared error (MSE) in decibels defined at a time step $n$ as $10 \log_{10} \frac{1}{T} \sum_{t=1}^{T} (e_{t,n})^2$ with $e_{t,n}$ being the error obtained by the current model of the considered method on the element $t$ of the testing set of size $T$.

\subsection{Simulation analysis}

\begin{figure}
    \centering
    \includegraphics[width=0.4\textwidth]{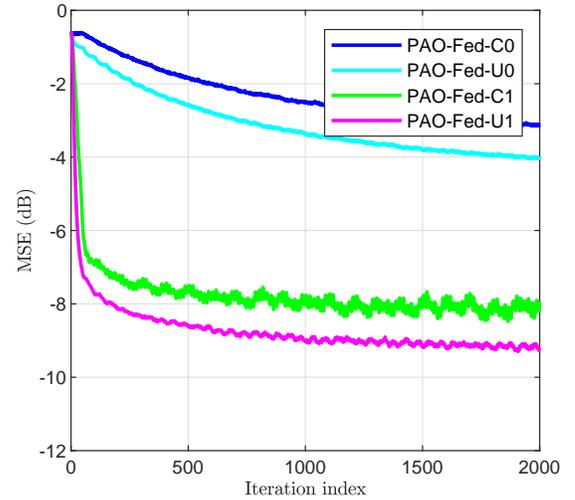}
    \caption{PAO-Fed performance on Setting I.}
    \label{Coordination}
\end{figure}

Fig. \ref{Coordination} shows the impact of the choice of the selection matrices on the convergence of the algorithm. First, we observe that taking advantage of the local updates performed by the unavailable clients receiving new data improve the convergence properties of the algorithm. For this reason, in the following, we will only consider the versions of PAO-Fed making full use of those updates. Second, we also observe that it is best to use uncoordinated partial-sharing in the presence of delays. This contradicts the behavior of partial-sharing FL in the absence of delays, where coordinated partial-sharing performs slightly better than uncoordinated \cite{Vinay}. The reason for this difference is that, in coordinated partial-sharing, all agents share the same subset of model parameters at a given global iteration. While in the absence of delays, this ensures that the aggregated server model parameters represent an average over a good number of clients, in the presence of delays, this means that the latest updates overwrite the previous ones, see \eqref{ServerUpdate}. For this reason, it is best not to set all the available clients to send the same subset of model parameters to the server.

\begin{figure}
    \centering
    \includegraphics[width=0.4\textwidth]{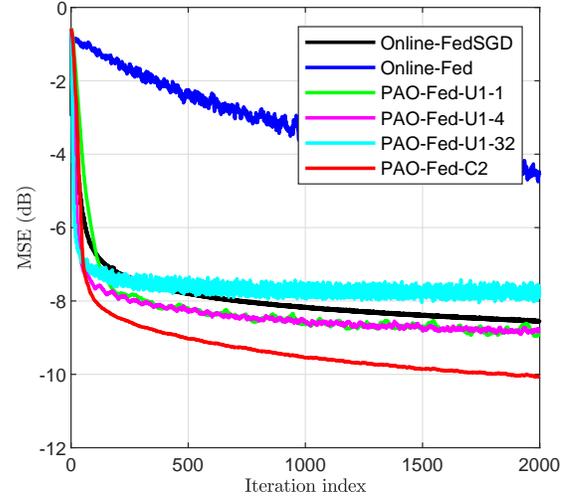}
    \caption{Choice of $m$ and $\alpha_l$ on Setting I.}
    \label{DelayWeight}
    \vspace{-3mm}
\end{figure}

Fig. \ref{DelayWeight} shows the learning curves for Online-FedSGD, Online-Fed, PAO-Fed-U1 for different values of $m$, and PAO-Fed-C2 that decreases the weight of delayed updates. We see that using local updates at the clients allows the PAO-Fed algorithm to achieves higher accuracy than Online-FedSGD while using $98\%$ less communication overhead. Second, we see that Online-Fed achieves poor accuracy in this setting; in fact, selecting a subset of clients in the already reduced pool of available participants is not a viable solution to reduce communication overhead in the asynchronous setting. Third, by comparing the PAO-Fed methods using $m = 1$, $4$, and $32$, we can see that, in the presence of delays, sharing more model parameters at each update do not necessarily increase accuracy as it does in a normal setting \cite{Vinay}. In fact, because it increases the potential negative impact of one single delayed update, it decreases accuracy. Sharing a small number of model parameters ensures better fitting of the server's model parameters to the overall average. Last, we observe that decreasing the weight of the received updates that have been delayed increases the accuracy of the algorithm. This is due to the fact that we prevent information coming from a small set of clients to overwrite the server's model parameters learned from a larger number of clients. Instead, we chose to consider the innovation coming from this delayed messages with reduced weight to take into account the lower relevance of these model parameters.

\begin{figure}
    \centering
    \includegraphics[width=0.4\textwidth]{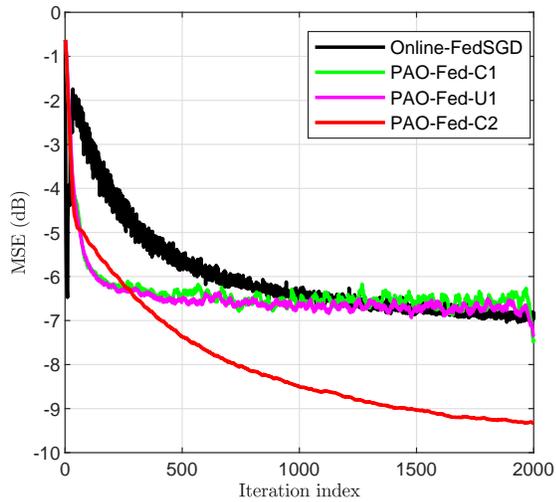}
    \caption{Performance on Setting II.}
    \label{LongDelay}
    \vspace{-5mm}
\end{figure}

In Figure \ref{LongDelay}, we observe the performance of the algorithms on Setting II, less favorable to learning. We observe that, in this setting, reducing the weight given to the delayed updates gains importance as the accuracy difference between PAO-Fed-C2 and PAO-Fed-U1 increases. This more significant difference is due to the fact that, in addition to overwriting the server's model parameters with model parameters from a smaller subset of clients, the delayed update provides some potentially outdated information.

\section{Conclusion}
We designed an energy-efficient FL algorithm adapted to a realistic environment. The proposed federated learning algorithm operates with significantly reduced communication requirements and can cope with an unevenly distributed system with poor client availability, channel blockage, and delays. Furthermore, the proposed partial-sharing mechanism reduces the communication overhead and diminishes the negative impact of delayed updates on accuracy. We further propose a weight-decreasing system for delayed updates that improve the performance of the algorithm, especially in an environment with poor participation and long delays.

\bibliographystyle{IEEEtran}
\bibliography{Federated}

\begin{thebibliography}{10}
\providecommand{\url}[1]{#1}
\csname url@samestyle\endcsname
\providecommand{\newblock}{\relax}
\providecommand{\bibinfo}[2]{#2}
\providecommand{\BIBentrySTDinterwordspacing}{\spaceskip=0pt\relax}
\providecommand{\BIBentryALTinterwordstretchfactor}{4}
\providecommand{\BIBentryALTinterwordspacing}{\spaceskip=\fontdimen2\font plus
\BIBentryALTinterwordstretchfactor\fontdimen3\font minus
  \fontdimen4\font\relax}
\providecommand{\BIBforeignlanguage}[2]{{%
\expandafter\ifx\csname l@#1\endcsname\relax
\typeout{** WARNING: IEEEtran.bst: No hyphenation pattern has been}%
\typeout{** loaded for the language `#1'. Using the pattern for}%
\typeout{** the default language instead.}%
\else
\language=\csname l@#1\endcsname
\fi
#2}}
\providecommand{\BIBdecl}{\relax}
\BIBdecl

\bibitem{origin}
J.~Kone{\v{c}}n{\`y}, H.~B. McMahan, D.~Ramage, and P.~Richt{\'a}rik,
  ``{Federated optimization: distributed machine learning for on-device
  intelligence},'' \emph{arXiv preprint arXiv:1610.02527}, Oct. 2016.

\bibitem{FedOverview}
T.~Li, A.~K. Sahu, A.~Talwalkar, and V.~Smith, ``{Federated learning:
  challenges, methods, and future directions},'' \emph{IEEE Signal Process.
  Mag.}, vol.~37, no.~3, pp. 50--60, May 2020.

\bibitem{Noniid}
Y.~Zhao, M.~Li, L.~Lai, N.~Suda, D.~Civin, and V.~Chandra, ``{Federated
  learning with non-iid data},'' \emph{arXiv preprint arXiv:1806.00582}, 2018.

\bibitem{FedAVG}
B.~McMahan, E.~Moore, D.~Ramage, S.~Hampson, and B.~A. y~Arcas,
  ``{Communication-efficient learning of deep networks from decentralized
  data},'' \emph{Artificial intel. statis.}, pp. 1273--1282, Apr. 2017.

\bibitem{Accelerated}
E.~Ozfatura, K.~Ozfatura, and D.~Gündüz, ``{FedADC: accelerated federated
  learning with drift control},'' in \emph{Proc. IEEE Int. Symp. Inf. Theory},
  Jul. 2021, pp. 467--472.

\bibitem{FedAvgCV}
X.~Li, K.~Huang, W.~Yang, S.~Wang, and Z.~Zhang, ``{On the convergence of
  fedavg on non-iid data},'' \emph{arXiv preprint arXiv:1907.02189}, Jul. 2019.

\bibitem{Compression}
J.~Kone{\v{c}}n{\`y}, H.~B. McMahan, F.~X. Yu, P.~Richt{\'a}rik, A.~T. Suresh,
  and D.~Bacon, ``{Federated learning: strategies for improving communication
  efficiency},'' \emph{arXiv preprint arXiv:1610.05492}, 2016.

\bibitem{Energy}
Z.~Yang, M.~Chen, W.~Saad, C.~S. Hong, and M.~Shikh-Bahaei, ``{Energy efficient
  federated learning over wireless communication networks},'' \emph{IEEE Trans.
  Wireless Commun.}, vol.~20, no.~3, pp. 1935--1949, Nov. 2020.

\bibitem{FLCompress1}
Z.~Lian, W.~Wang, and C.~Su, ``{COFEL: Communication-efficient and optimized
  federated learning with local differential privacy},'' in \emph{Proc. IEEE
  Int. Conf. Commun.}, Jun. 2021, pp. 1--6.

\bibitem{FLCompress2}
Y.~Lu, Z.~Liu, and Y.~Huang, ``{Parameters compressed mechanism in federated
  learning for edge computing},'' in \emph{Proc. IEEE Int. Conf. Cyber Secur.
  Cloud Comput.}, Jun. 2021, pp. 161--166.

\bibitem{FLCompress3}
X.~Fan, Y.~Wang, Y.~Huo, and Z.~Tian, ``{Communication-efficient federated
  learning through 1-bit compressive sensing and analog aggregation},'' in
  \emph{Proc. IEEE Int. Conf. Commun. Workshops}, 2021, pp. 1--6.

\bibitem{Async}
Y.~Chen, Z.~Chai, Y.~Cheng, and H.~Rangwala, ``{Asynchronous federated learning
  for sensor data with concept drift},'' \emph{arXiv preprint
  arXiv:2109.00151}, Sep. 2021.

\bibitem{Async2}
C.~Xie, S.~Koyejo, and I.~Gupta, ``{Asynchronous federated optimization},''
  \emph{arXiv preprint arXiv:1903.03934}, Mar. 2019.

\bibitem{AsyncOnline}
Y.~Chen, Y.~Ning, M.~Slawski, and H.~Rangwala, ``{Asynchronous online federated
  learning for edge devices with non-iid data},'' in \emph{Proc. IEEE Int.
  Conf. Big Data}, Dec. 2020, pp. 15--24.

\bibitem{AsyncTier}
Z.~Chai, Y.~Chen, L.~Zhao, Y.~Cheng, and H.~Rangwala, ``{Fedat: a
  communication-efficient federated learning method with asynchronous tiers
  under non-iid data},'' \emph{arXiv preprint arXiv:2010.05958}, Oct. 2020.

\bibitem{CommAsync}
Y.~Chen, X.~Sun, and Y.~Jin, ``{Communication-efficient federated deep learning
  with layerwise asynchronous model update and temporally weighted
  aggregation},'' \emph{IEEE Trans. Neural Netw. Learn. Syst.}, vol.~31,
  no.~10, pp. 4229--4238, Dec. 2019.

\bibitem{Asyncstuf}
Z.~Wang, Z.~Zhang, and J.~Wang, ``{Asynchronous federated learning over
  wireless communication networks},'' in \emph{Proc. IEEE Int. Conf. Commun.},
  Jun. 2021, pp. 1--7.

\bibitem{Vinay}
V.~C. Gogineni, S.~Werner, Y.-F. Huang, and A.~Kuh, ``{Communication-efficient
  online federated learning framework for nonlinear regression},''
  \emph{https://arxiv.org/abs/2110.06556}, Oct. 2021.

\bibitem{DblCompress}
``{Robust and communication-efficient federated learning from non-i.i.d.
  data},'' \emph{IEEE Trans. Neural Netw. Learn. Syst.}, vol.~31, no.~9, pp.
  3400--3413, Sep. 2020.

\bibitem{RandomMask}
J.~Kone{\v{c}}n{\`y}, H.~B. McMahan, F.~X. Yu, P.~Richt{\'a}rik, A.~T. Suresh,
  and D.~Bacon, ``{Federated learning: Strategies for improving communication
  efficiency},'' \emph{arXiv preprint arXiv:1610.05492}, Oct. 2016.

\bibitem{PartialSharing}
R.~Arablouei, K.~Doğançay, S.~Werner, and Y.-F. Huang, ``{Adaptive
  distributed estimation based on recursive least-squares and partial
  diffusion},'' \emph{IEEE Trans. Signal Process.}, vol.~62, no.~14, pp.
  3510--3522, Jul. 2014.

\bibitem{RFFKLMSB}
A.~Rahimi, B.~Recht \emph{et~al.}, ``{Random features for large-scale kernel
  machines.}'' in \emph{Proc. Conf. on Neural Inf. Proc. Syst.}, vol.~3, no.~4,
  Dec. 2007, pp. 1--5.

\bibitem{RFFKLMS}
P.~Bouboulis, S.~Pougkakiotis, and S.~Theodoridis, ``{Efficient KLMS and KRLS
  algorithms: a random Fourier feature perspective},'' in \emph{Proc. IEEE
  Stat. Signal Process. Workshop}, Jun. 2016, pp. 1--5.

\bibitem{AoI}
R.~D. Yates, Y.~Sun, D.~R. Brown, S.~K. Kaul, E.~Modiano, and S.~Ulukus, ``{Age
  of information: an introduction and survey},'' \emph{IEEE J. Sel. Areas
  Commun.}, vol.~39, no.~5, pp. 1183--1210, Mar. 2021.

\bibitem{tcas2}
V.~C. Gogineni, S.~P. Talebi, and S.~Werner, ``Performance of clustered
  multitask diffusion lms suffering from inter-node communication delays,''
  \emph{IEEE Trans. on Circuits and Syst. II: Express Briefs}, vol.~68, no.~7,
  pp. 2695--2699, 2021.

\end{thebibliography}

\end{document}